\documentclass[10pt,twocolumn,letterpaper]{article}

\usepackage{iccv}
\usepackage{times}
\usepackage{epsfig}
\usepackage{graphicx}
\usepackage{amsmath}
\usepackage{amssymb}

\usepackage[bottom]{footmisc}

\usepackage{placeins}
\usepackage{booktabs}
\usepackage{array}
\usepackage[font=small,format=plain,labelfont=bf,up,textfont=normal,up,justification=justified,singlelinecheck=false]{caption}

\newcommand{\at}{\makeatletter @\makeatother}

\newcommand{\secref}[1]{Sect.~\ref{#1}}
\newcommand{\figref}[1]{Fig.~\ref{#1}}
\newcommand{\tabref}[1]{Table~\ref{#1}}

\usepackage[breaklinks=true,bookmarks=false]{hyperref}

\iccvfinalcopy % *** Uncomment this line for the final submission

\def\iccvPaperID{1817} % *** Enter the ICCV Paper ID here

\ificcvfinal\fi
\begin{document}

%%%%%%%%% TITLE
\title{Self-supervised Learning of Pose Embeddings \\ from Spatiotemporal Relations in Videos}

\author{\"{O}mer S\"{u}mer\thanks{Both authors contributed equally to this work.} \qquad Tobias Dencker\footnotemark[1]  \qquad  Bj\"{o}rn Ommer\\
Heidelberg Collaboratory for Image Processing\\
IWR, Heidelberg University, Germany\\
{\tt\small firstname.lastname@iwr.uni-heidelberg.de}
}

\maketitle
\thispagestyle{empty}

%%%%%%%%% ABSTRACT
\begin{abstract}
	Human pose analysis is presently dominated by deep convolutional networks trained with extensive manual annotations of joint locations and beyond. To avoid the need for expensive labeling, we exploit spatiotemporal relations in training videos for self-supervised learning of pose embeddings. The key idea is to combine temporal ordering and spatial placement estimation as auxiliary tasks for learning pose similarities in a Siamese convolutional network. Since the self-supervised sampling of both tasks from natural videos can result in ambiguous and incorrect training labels, our method employs a curriculum learning idea that starts training with the most reliable data samples and gradually increases the difficulty. To further refine the training process we mine repetitive poses in individual videos which provide reliable labels while removing inconsistencies. Our pose embeddings capture visual characteristics of human pose that can boost existing supervised representations in human pose estimation and retrieval. We report quantitative and qualitative results on these tasks in Olympic Sports, Leeds Pose Sports and MPII Human Pose datasets.
\end{abstract}

\section{Introduction}
The ability to recognize human posture is essential for describing actions and comes natural to a human being. Different poses in a video form a visual vocabulary similar to words in text. An important objective of computer vision is to bring this ability to the computer. Finding similar postures across different videos automatically enables a lot of different applications like action recognition \cite{Antic2014,Antic2015} or video content retrieval.

So what makes two postures look similar? More formally, a similarity function, which is entailed by a pose embedding, needs to capture the characteristics of different postures, while exhibiting the necessary invariance to strong intra-class variations. In particular, it should be sensitive to articulation of body parts while being invariant to illumination, background, clutter, deformations (e.g. facial expressions) or occlusions. Often human joints are used as a surrogate for describing similarity, but there are several issues: First, measuring distances in pose space accurately and coming up with a non-ambiguous Euclidean embedding is already a challenging problem. Second, the manual annotation of human joints in larger datasets is expensive and time-consuming.

Convolutional networks have recently been immensely helpful to computer vision. The feature hierarchy of such a network is effectively defined by a cascade of filter banks that are recursively applied to extract discriminative features for the given task. In this work we take advantage of convolutional networks to learn pose embeddings from videos. 

In supervised similarity learning we assume that we are given positive and negative pairs of postures for training. In this supervised setting convolutional networks excel and have recently surpassed human performance in some basic tasks. In contrast unsupervised training of convolutional networks is still an open problem and currently the focus of the research community. In this paper we investigate how to learn a pose representation without labels.

A solution for the problem of missing supervision is to switch to a related auxiliary task for which label information is available. For this self-supervised strategy several well-known sources of weak supervision have been recently re-visited: among them spatial configuration of natural scenes, inpainting, super-resolution, image colorization, tracking, ego-motion and even audio. Although there are many sources available, not all of them are suitable for the application in pose analysis. We exploit human motion in videos to make pose similarity apparent and learnable without labels. With an almost infinite supply of video data online exploiting this idea is very attractive.

We propose learning spatiotemporal relations in videos by means of two complementary auxiliary tasks: a \emph{temporal ordering} task which learns whether two given person images are temporally close (similar) and a \emph{spatial placement} task which discovers randomly extracted crops from the spatial neighborhood of persons, and learns whether given patches are a person or not. Learning spatial and temporal relations of human movement simultaneously provides us information of ``what'' we are looking at (person/ not person) and ``how'' the instances differ (similar/dissimilar poses). \emph{Curriculum-based} learning and \emph{repetition mining} arrange the training set by starting from only the easy samples and then iteratively extend to harder ones, while also eliminating inactive video parts. Then our spatiotemporal embeddings successfully learn representative features of human pose in a self-supervised manner.

\begin{figure*}[htp!]
	\centering
	%\fbox{\rule{0pt}{2in} \rule{.9\linewidth}{0pt}}
	\includegraphics[width=.81\linewidth]{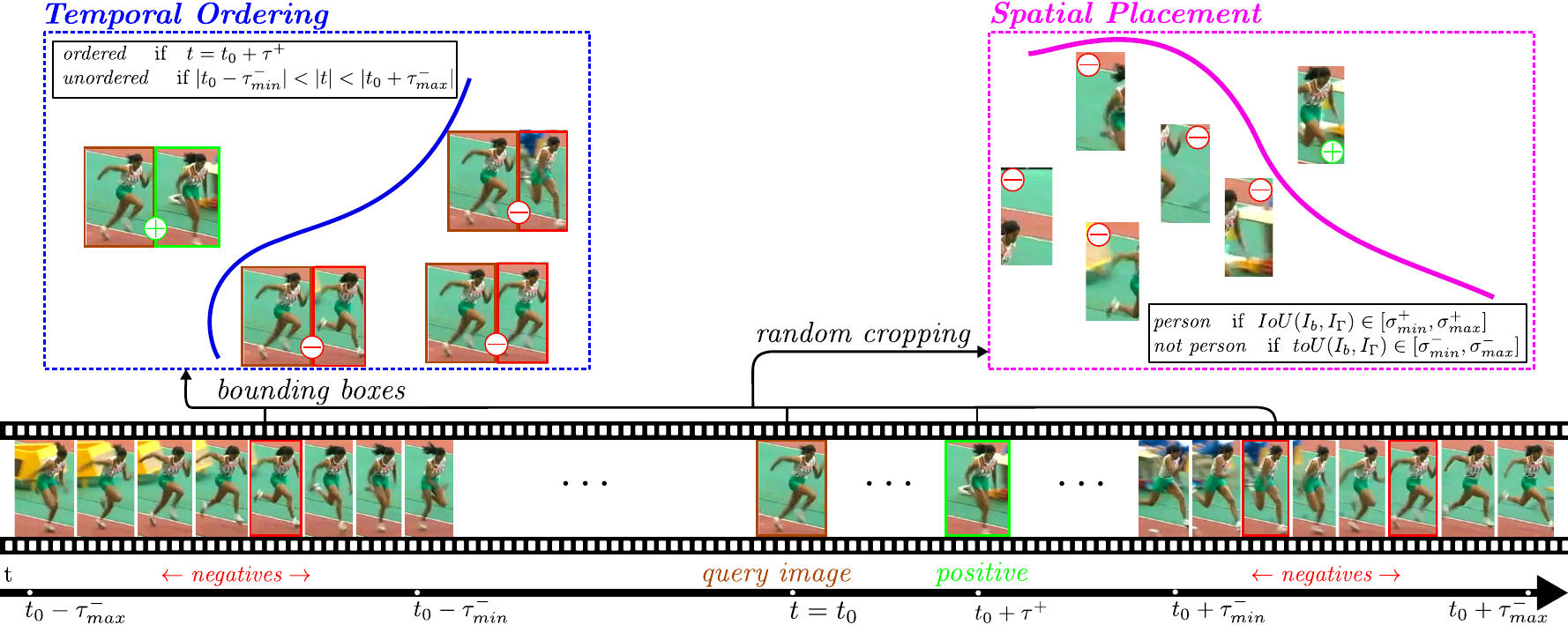}
	\caption{Sampling procedure for training self-supervised pose embeddings. For each query image in a video, positive and negative pairs of temporal ordering are collected from specific temporal ranges \emph{(left)}. In spatial placement, samples are cropped using the IoU criterion \emph{(right)}.}
    % Self-supervised pose embeddings: temporal ordering and spatial placement. Example video from Olympic Sports dataset \cite{Niebles2010} shows how our auxiliary tasks create training data.
	\label{fig:auxiliary_tasks}
\end{figure*}

\section{Related Work}
Human pose analysis deals with problems such as pose retrieval, similarity learning and pose estimation. Most approaches in pose analysis rely on supervised data and there exists only a few unsupervised approaches. Here, we summarize significant examples of pose analysis and related unsupervised learning approaches:\par

\textbf{Pose estimation} \ \ Pose estimation aims at finding locations of body joints, whereas pose retrieval or embedding finds a metric that can retrieve the most similar poses and discriminates samples according to their pose information, without localizing joints directly. With the advancements in convolutional neural networks \cite{Krizhevsky_NIPS2012}, pose estimation is also dominated by deep learning-based methods. Toshev and Szegedy \cite{Toshev_CVPR2014} estimated joint locations directly regressing in a CNN architecture. Instead of simply regressing joint locations, Chen and Yuille \cite{Chen_NIPS2014} learned pairwise part relations combining CNN with graphical models. Tompson \emph{et al.} \cite{Tompson_NIPS2014} exploited CNNs for relationship between body parts with a cascade refinement. A recent work by Newell \emph{et al.} \cite{Newell_ECCV2016} used fully convolutional networks in a bottom-up top-down manner to predict heatmaps for joint locations. \par

\textbf{Similarity learning} \ \ The first \emph{Siamese}-type architecture \cite{Bromley_NIPS1994} was proposed to learn a similarity metric for signature verification. Similarity learning was also applied in human pose analysis. In \cite{Mori_ArXiv2015} and \cite{Kwak_CVPR2016}, body joint locations are used to create similar and dissimilar pairs of instances from annotated human pose datasets. \cite{Kwak_CVPR2016} also transferred a learned pose embedding to action recognition.\par

These works in pose estimation and similarity learning exploited large amounts of annotations (body joints or labeling of similar/dissimilar postures). However, unsupervised learning methods without using labels showed promising performance in various learning tasks in the last decade. Self-supervised learning is very popular similar to classical unsupervised methods such as clustering, autoencoders \cite{Rumelhart1986}, restricted Boltzman machines \cite{Hinton1983}. % Rumelhart1986,Hinton1986

\textbf{Self-supervised learning} \ 
The availability of big data motivated the community to investigate alternative sources of supervision such as ego-motion \cite{Agrawal2015,Zamir2016}, colorization \cite{Zhang2016}, image generation \cite{Radford2015}, spatial \cite{Doersch_ICCV2015,Norozzi_ECCV2016} or temporal clues \cite{Wang_ICC2015,Misra_ECCV2016}. As our approach belongs to the class of self-supervised methods using spatial and temporal information, we describe these methods in detail.

Doersch \emph{et al.} \cite{Doersch_ICCV2015} trained convolutional neural networks to take image patches from a $3\times3$ grid and classify the relative location of 8 patches with respect to a center patch. Norozzi and Favaro \cite{Norozzi_ECCV2016} argued that solving locations of relative patches could introduce ambiguities and proposed a localization problem given all 9 patches at once. Also, they used 100 relative locations as class labels out of $9!$ permutations using a Hamming distance-based selection. 

Wang and Gupta \cite{Wang_ICC2015} exploited videos by detecting interesting regions with SURF keypoints and tracking them. Then, they used a Siamese-triplet architecture with a ranking loss together with random negative selection and hard negative mining. However, tracking is not the best solution in the challenging context of pose analysis due to the non-rigid deformations of person patches which are in low resolution and contain too few keypoints to detect parts and track them precisely.

Misra \emph{et al.} \cite{Misra_ECCV2016} defined a temporal order verification task, which classifies whether given 3-frame sequences are temporally ordered or not by altering the middle frame. In action/pose benchmarks or internet videos, there are a lot of cyclic human actions (\emph{e.g.} running based sports, dancing), which often produce confusing samples and interfere with representation learning.  

In order to learn a better representation, we argue that temporal cues which aim to learn whether given inputs are from temporally close windows or not will be a more effective approach. The use of temporal cues to learn whether given inputs are from temporally close windows or not is an effective approach for representation learning. Local proximity in data (slow feature analysis, SFA) has first been proposed by Becker and Hinton \cite{Becker1992}. The most recent spatial and temporal self-supervised learning methods are inspired from SFA. Goroshin \emph{et al.} \cite{Goroshin2015} created a connection between slowness and metric learning by temporal coherence. Motivated by temporal smoothness in feature space, Jayaraman and Grauman \cite{Jayaraman2016} exploited higher order coherence, which they referred to as steadiness, in various tasks. Slowness or steadiness criterion can introduce significant drawbacks mostly because of limited motion and the repetitive nature of human actions. Thus, we learn auxiliary tasks in relatively small temporal windows which do not contain more than a single cycle of action. Moreover, the use of curriculum learning \cite{Bengio2009} and repetition mining refine and guide our self-supervised tasks to learn stronger temporal features.   

Curriculum learning has been proposed by Bengio \emph{et al.} \cite{Bengio2009} and it speeds up training and improves test performance by using samples whose difficulties are gradually increasing in shape recognition and language modeling. To the best of our knowledge, the potential of a curriculum has not been studied in the self-supervised setting, where we associate the difficulty of training samples with their inherent motion.

%In comparison with related self-supervised works \cite{Doersch_ICCV2015,Norozzi_ECCV2016,Wang_ICC2015,Misra_ECCV2016} our method leverages two prominent cues for learning pose representations: \emph{temporal ordering} and \emph{spatial placement} jointly. Our temporal ordering task eliminates possible confusion due to challenging spatiotemporal dynamics of human motion (repetitiveness, occlusion, change in point of view, camera panning etc.) by use of a new optical flow-based curriculum and mining repeated poses with the learned representation in self-supervised manner. As a result of these refinements, our embeddings learn the characteristics of posture efficiently and even generalize to different datasets in pose analysis without transfer learning.  

\section{Approach} \label{section:approach}

Our motivation is to learn pose embeddings from videos without labels. We follow the insight that spatiotemporal relations in videos provide sufficient information for learning. For this purpose, we propose a self-supervised pipeline that creates training data for two auxiliary tasks: 1) temporal ordering and 2) spatial placement. Since the raw self-supervised output needs refinement, we introduce curriculum learning and repetition mining as key ingredients for successful learning. The two auxiliary tasks are trained in a Siamese CNN architecture and the learned features are eventually used as pose embeddings in order to retrieve similar postures and estimate pose.

\subsection{Self-supervised Pose Embeddings: Temporal Ordering and Spatial Placement} \label{subsection:auxiliarytasks}

We consider a temporal and a spatial auxiliary task which are automatically sampled from videos as described in \figref{fig:auxiliary_tasks}. Both tasks capture complementary information from inside videos essential for learning a pose embedding. The temporal task teaches the pose embedding to become more sensitive to body movements and more invariant to camera motion (\emph{i.e.} panning, zoom in/out, jittering), while the spatial task relies on the spatial configuration of a single frame and focuses on learning a human appearance model which strengthens the ability to separate posture from background. 

For the temporal ordering task, a tuple of two frames is sampled from the same video together with a binary label which indicates whether the first frame (anchor) is closely followed in time by the second frame (candidate). In order to focus on learning human posture, we do not sample the full frames, but instead crop bounding box estimates of the person of interest. Thus, the training input for the temporal ordering task consists of two cropped boxes and a binary label indicating whether the two boxes are temporally ordered.

For a frame $I_{t_0}$ sampled at time point $t_0$, we sample a candidate frame $I_{t}$ with a temporal offset of $\Delta_t=t-t_0$. In order to sample a positive candidate the offset needs to be $\Delta_t= \tau^+$, while a negative candidate is sampled if 
\[\Delta_t \in \tau^- = [\tau_{min}^-, \tau_{max}^-] \cup [-\tau_{max}^-, -\tau_{min}^-]\]
%$\Delta_t \in \tau^- = [\tau_{min}^-, \tau_{max}^-] \cup [-\tau_{max}^-, -\tau_{min}^-]$
holds. $\tau_{min}^-, \tau_{max}^-$ are the range limits of the negative candidates. In other words, a positive candidate comes exactly from $\tau^+$ frames in the future, while negative candidates come from ranges before or after the anchor frame. 

The temporal ordering task relies on the assumption of temporal coherence that frames in a small temporal neighborhood are more similar than distant frames. We add the constraint that positive candidates can only come from the future. Since the self-supervised sampling from videos already introduces large amounts of variation, we want the positive class to be as homogeneous as possible in order to facilitate training. In contrast the negative class is sampled from a larger range that allows more variation, but is still close enough to the positive class to provide challenging similarities for discriminative learning.

For the spatial placement task, a box is randomly cropped from a single frame together with a binary label that indicates whether the cropped box overlaps with the estimated bounding box of a person in this frame. The overlap is measured with the Intersection-over-Union (IoU) criterion \cite{Everingham2010}. For the estimated bounding box $I_b$ and a randomly cropped box $I_r$, the binary label $y_S$ is defined as
\begin{equation}
y_S(I_b, I_r)= 
\begin{cases}
1, 		& \text{if } \ IoU(I_b, I_r) \in [\sigma^+_{min},\sigma^+_{max}] \\
0,      & \text{if } \ IoU(I_b, I_r) \in [\sigma^-_{min},\sigma^-_{max}]
\end{cases} \; 
\end{equation}
where $IoU(\cdot , \cdot)$ computes the IoU and $[\sigma^+_{min},\sigma^+_{max}]$ defines the positive range of overlap while $[\sigma^-_{min},\sigma^-_{max}]$ defines the negative. Since the estimated bounding boxes are not completely reliable, the positive and negative IoU ranges are usually selected with a gap between them to help the separation of the classes.

In both auxiliary tasks, three negative samples are used for each positive posture, because sampling of negatives (what it is not) from larger ranges helps with learning positive similarities (what it is) precisely. The intuition is that the pose embedding learns to discriminate between a homogeneous positive and a more heterogeneous negative class in both tasks. Since both tasks focus on different aspects of human posture, the best pose embedding is obtained by joint training. We investigate the contribution of different configurations in \secref{subsection:postureanalysis}. 

\begin{figure*}[ht!]
	\begin{center}
		%\fbox{\rule{0pt}{2in} \rule{.9\linewidth}{0pt}}
		\includegraphics[width=.85\linewidth]{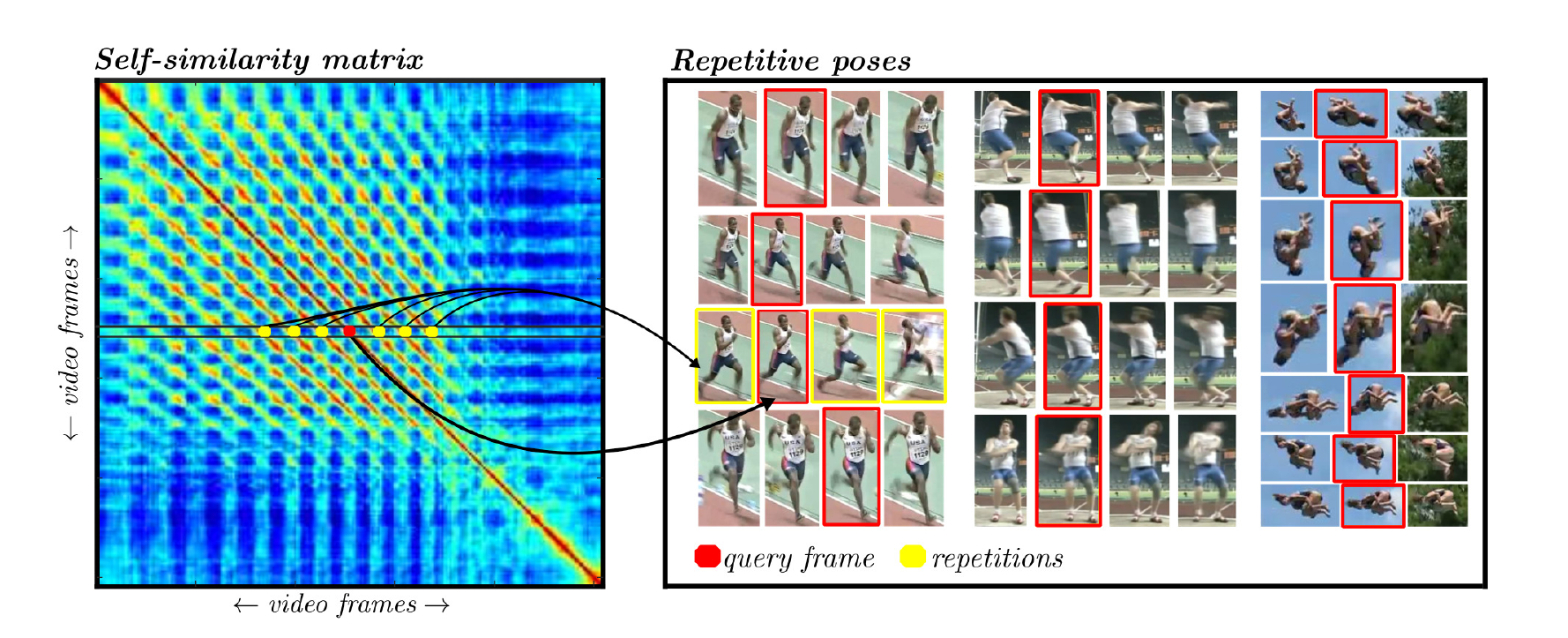}
	\end{center}
	\caption{Mining repetitive poses. Off-diagonal structures of the self-similarity matrix on the left indicate repetitions in a video. For each row, repetitions are mined using a query frame. Repetitive poses from three videos are shown on the right.}
    % Mining repetitive poses. An example self-similarity matrix \emph{(left)} and repetitive poses \emph{(right)}.
	\label{fig:repetitions}
\end{figure*}

\subsection{Creating a Curriculum for Training} \label{subsection:curriculum}

In supervised training with human annotations, it is often beneficial to avoid difficult samples with ambiguous or even incorrect labels, because this kind of data can inhibit convergence and lead to inferior results. In the self-supervised case, we find that data quality fluctuates even more and needs to be taken into account. On the other hand, skipping too many difficult training samples can result in overfitting on a small subset of easy samples and hurts generalization to unseen datasets. We propose to strike a balance by using a curriculum of training data that gradually increases in difficulty over the course of training. We create the curriculum with regard to the temporal ordering task which produces far more inconsistent samples than spatial placement. 

In order to determine the difficulty of temporal ordering for a particular training sample, we look into the motion characteristics of the respective video. For instance, a clean-and-jerk video mainly consists of inactive parts with little motion, whereas a long-jump video is dominated by a highly repetitive structure with fast moving, deforming postures. Training samples from video sequences with clear foreground motion (e.g.\ a long-jump video) are preferable for learning temporal ordering, because their negative candidates, which are sampled from the range of $\tau^-$, are easier to distinguish from the positive ones from $\tau^+$. Therefore, we determine the difficulty of a training sample by estimating the motion in videos and sample training frames with sufficient action. 

When creating a curriculum, we use an optical flow based criterion that computes the ratio of the optical flow in the foreground and background of the frame. To compute the \emph{fg/bg ratio} the mean magnitude of optical flow in the foreground bounding box is divided by mean magnitude of optical flow of the background. We use the method from \cite{Brox2004} to estimate the optical flow between two frames. The fg/bg ratio acts as a proxy of a \emph{signal-to-noise} ratio, as examples with higher values are more easily separated from the background. 

The curriculum is assembled by sorting the training samples according to their flow ratio and splitting them in discrete blocks, curriculum updates, with increasing difficulty (decreasing flow ratio). We analyze the impact of the curriculum in an ablation experiment in \tabref{table:postureOSD1} where we train the network with and without a curriculum using the same subset of self-supervised training data. Details of ablation experiments and the effect of curriculum will be explained in \secref{subsection:trainingsetup} and \secref{subsection:postureanalysis}.

\subsection{Mining Repetitive Poses} \label{subsection:repetitions}

There are two reasons why we pay special attention to repetitive poses in video sequences: First, they impair the training of the temporal ordering task. Second, if the location of repetitions were known, they could be extracted and used as valuable training data, which we refer to as \emph{repetition mining}. The mined repetitions augment temporal ordering by providing a new similarity learning task.

While the proposed curriculum avoids difficult samples in the early stages of self-supervised training, repetitive poses in videos are not filtered by the motion-based curriculum. The training of the temporal ordering task suffers from repetitions which can cause incorrect labeled image pairs by violating the assumption of temporal coherence. For instance, if a negative frame is sampled from a video with a repetitive action like running or walking, it might be more similar to the anchor frame than the positive candidate. 

After an initial training of the temporal ordering task, we use the learned pose embeddings to detect repetitive poses in the training data. For each video, we obtain a self-similarity matrices by computing all the pairwise distances between frames. As distance measure, we use the Euclidean norm of the normalized \texttt{pool5} features. In order to extract reliable and strong repetitions, we convolve the self-similarity matrix with a 5x5 circulant filter matrix to suppress potential outliers that are not aligned with the off-diagonals by thresholding. The maxima of each row indicate the fine-scaled repetitions of the respective query frame. \figref{fig:repetitions} shows an example self-similarity matrix and repetitions which are mined using this approach.

Repetitive poses form groups of very similar but not identical images due to small variations over time caused by the persons movement, changes in the camera viewpoint, or even the frame rate of the video camera. These groups of highly similar images help to learn the more fine-grained details of human posture. They can be used to create a new type of similar-dissimilar problem. Similar pairs are chosen among repetition groups, negative candidates are picked from regions between the repetitions. 

As repetitions occur only in a subset of the available video data, they are combined with samples from non-repetitive videos and added to the first stages of the curriculum. The mined repetitive poses are in quality close to human annotated similarities and provide a stabilizing effect on the whole training procedure. 

%Therefore our curriculum guarantees that during training at least a quarter of training samples are repetitive poses.

Our method can be employed in a bootstrapping fashion, by repeatedly training the temporal ordering task and mining repetitions which provide better training samples without additional supervision.

\subsection{Network Architecture} \label{subsection:network}
\begin{figure}[b]
	\begin{center}
		%\fbox{\rule{0pt}{2in} \rule{.9\linewidth}{0pt}}
		\includegraphics[width=.99\linewidth]{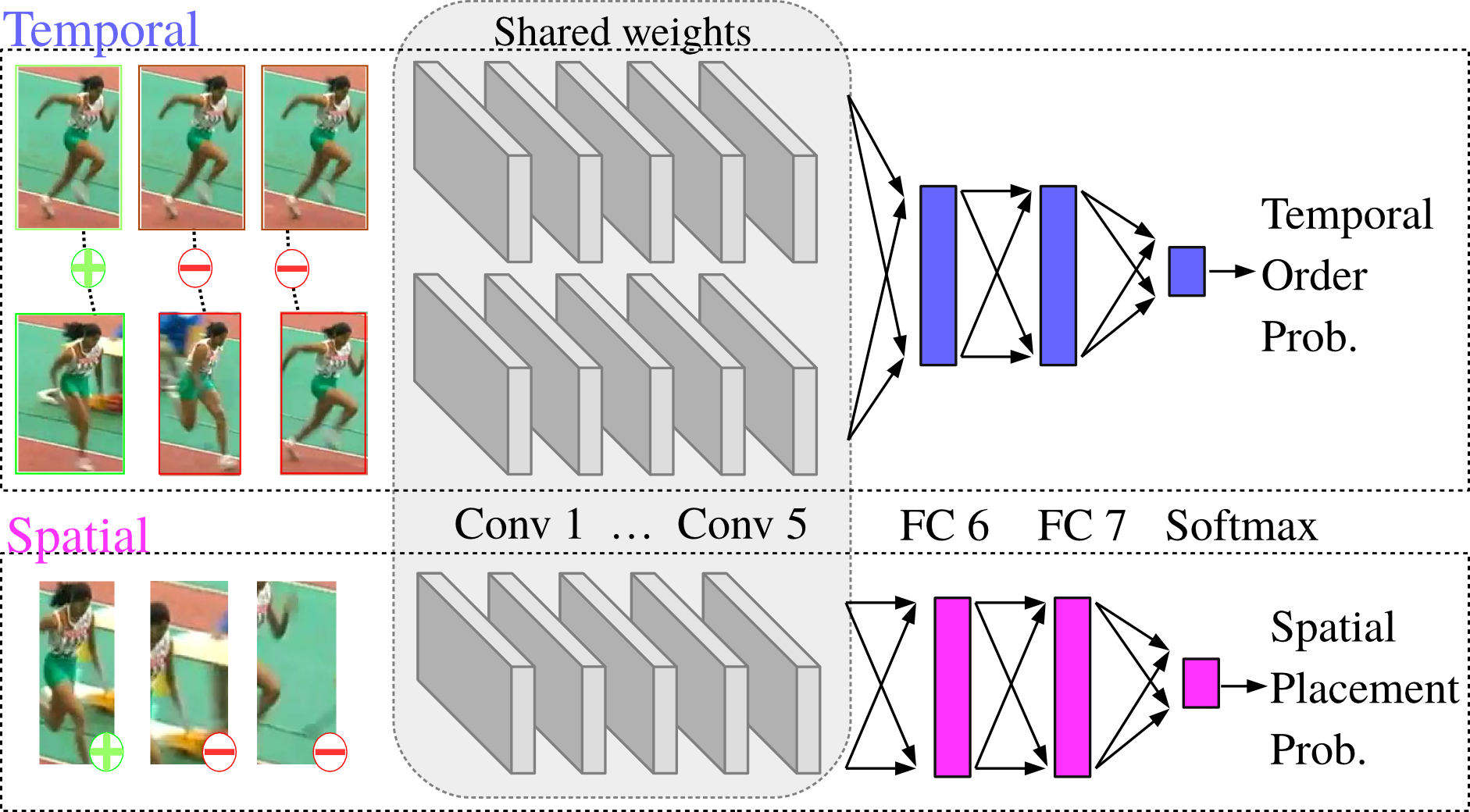}
	\end{center}
	\caption{Network architecture for temporal ordering and spatial placement.}
	%The temporal ordering feeds pairs of images as input, while the spatial placement task takes image crops. First five convolutional layers share parameters. Fully connected layers have 2048 and 1024 neurons and are regularized with dropout and batch normalization. Both networks produce a binary output using a softmax function.
	\label{fig:network}
\end{figure}
For the two self-supervised tasks we train two convolutional neural networks which differ in the number of images they process as shown in \figref{fig:network}. The temporal ordering task is trained using a Siamese architecture \cite{Bromley_NIPS1994} that takes a pair of images as input while the spatial placement task is trained on single images using a common single stream architecture.

We adopt the well-known Alexnet architecture \cite{Krizhevsky_NIPS2012} for both tasks. In the temporal task the two Siamese streams consist of convolutional layers. After the last pooling layer the output from the two streams is concatenated. The fully-connected layers compute a binary output probability for testing. The convolutional networks are trained by minimizing binary cross-entropy loss functions. For joint training of both tasks the weights in the convolutional layers are not only shared between the Siamese parts but are also shared with the convolutional layers of the spatial placement task. Moreover, the joint loss of the two auxiliary tasks is computed in a weighted sum.

After training the network, we use the feature representation from the last shared layer \texttt{Pool5} as pose embeddings. Features of this layer provide good localization which is important for pose retrieval and estimation.

We make several modifications to the Alexnet architecture: 1) Because we want to avoid overfitting and our binary tasks do not require a large number of parameters, both networks have a reduced number of neurons in the fully connected layers compared to the original Alexnet (namely 2048/1024 vs 4096/4096). 2) To improve training of the temporal task we replace the regular rectified linear unit in the last convolutional layer with a non-linearity that has a negative slope. We find that this modification is critical for performance. 3) The use of batch normalization in the fully connected layers is an important regularizer in our training that helps with generalization to other datasets.

\section{Experiments} \label{section:experiments}
We present experiments on posture analysis, pose estimation and pose retrieval. The training of our method is demonstrated on the Olympic Sports dataset (OSD) \cite{Niebles2010}. In different ablation experiments we highlight the design decisions in our proposed method. To study the ability of our approach to generalize to unseen datasets we include experiments on Leeds Sports Pose (LSP) \cite{Johnson2010} and the challenging and unconstrained MPII Human Pose \cite{Andriluka2014}. Additionally in a supervised pose estimation setting \cite{Toshev_CVPR2014}, we report performance of our method in comparison with other initialization approaches.

\subsection{Training and Testing Details} \label{subsection:trainingsetup}

From 680 videos in Olympic Sports dataset, we extract approximately 140,000 frames for which we obtain bounding box estimates using the method in \cite{Felzenszwalb2010}. Our training curriculum uses about 80,000 frames which are ordered using the flow ratio criterion described in \secref{subsection:curriculum}. It starts out with about five percent of the easiest training samples and increases the amount of training data in seven steps every 2.5K iterations. The amount of training data grows exponentially during the first few curriculum updates, but does not surpass 25 percent of training data for a single update.

For training of the convolutional networks we use the \emph{Caffe} framework \cite{Jia2014}. We optimize our model using the Adam solver for stochastic batch gradient descent with batch size of 48 and a fixed learning rate of $10^{-4}$.\ In the convolutional layers, we use a reduced learning rate of $10^{-5}$.\ The training is stopped after 40K iterations. For joint training we reduce the loss weight for the spatial task by a factor of 0.1. In the auxiliary tasks we use $\tau^+=4$ and $\tau^-=[8,16]$ as well as $\sigma^+=[0.65,0.95]$ and $\sigma^-=[0.25,0.55]$. For mining repetitions we follow the procedure described in \secref{subsection:repetitions} and iterate it two times to collect about 15000 frames with repetitive poses. We find that two iterations are sufficient, since our method has found most of the repetitions by this time. For testing, we use the pairwise Euclidean distance of \texttt{Pool5} features as a similarity measure between images.

\subsection{Ablation Experiments on Posture Analysis in Olympic Sports Dataset} \label{subsection:postureanalysis}

We demonstrate on the Olympic Sports dataset, how different configurations of our method affect the performance of the learned pose embeddings. For the evaluation on the the Olympic Sports dataset, we adopt the posture analysis benchmark proposed in \cite{Bautista2016}. It consists of 1200 exemplar postures for each of which ten positive (similar) and ten negative (dissimilar) poses are defined. The performance is determined by the ability of the pose embeddings to separate positives from negatives and measured in terms of the area under the curve (AuC) of a ROC. 

\begin{table}[tb]
	\centering
	\resizebox{\columnwidth}{!}{
		
		\begin{tabular}{*3l} \toprule
			Task & without curriculum & with curriculum \\\midrule
			temporal(T) & 0.592 & 0.630 \\ 
			temporal\& spatial & 0.664 & 0.679 \\\midrule 
			temporal(T)$\ast$ & 0.762 & 0.781 \\
			temporal\& spatial$\ast$ & 0.767 & 0.784  \\\bottomrule
		\end{tabular}
		
	}
	\caption{Average AUC in Olympic Sports benchmark shows effect of curriculum training. Methods with $(\ast)$ are initialized with Imagenet pre-trained weights.}
	\label{table:postureOSD1}
\end{table} 

First, we study the impact of curriculum learning. We train our temporal (T) and temporal \& spatial (ST) tasks once with and once without a curriculum, but using the same amount of training data. The experiments in \tabref{table:postureOSD1} show that the curriculum as proposed in \secref{subsection:curriculum} improves the performance of our method by 5\% in mean AUC in random initialized temporal task. When the temporal task is initialized with Imagenet pre-trained weights, it improves by 2\%, and this improvement is preserved even in joint learning of temporal ordering and spatial placement. Temporal ordering itself seems less powerful than spatial placement and cannot be learned without curriculum learning. However, our \emph{fg/bg ratio} based curriculum significantly increases its performance in posture analysis.

\begin{table}[tb] % bp
	\centering
	%\resizebox{0.9\columnwidth}{!}{
	\begin{tabular}{*3l} \toprule
		Method & Avg. AUC \\\midrule
		temporal(T) & 0.630 \\
		spatial(S)  & 0.668 \\ 
		temporal \& spatial & 0.679\\
		T with repetitions & 0.658\\
		S\&T with repetitions & \textbf{0.701} \\\midrule
		HOG-LDA 							 				& 0.580\\
		Doersch \emph{et al.} \cite{Doersch_ICCV2015} & 0.580\\
		Jigsaw puzzles \cite{Norozzi_ECCV2016} (Imagenet)  &  0.653\\
		Jigsaw puzzles \cite{Norozzi_ECCV2016} (OSD)       &  0.646\\
		Shuffle\&Learn \cite{Misra_ECCV2016}               &  0.646\\
		Video triplet  \cite{Wang_ICC2015}                 &  0.598\\\midrule
		Alexnet \cite{Krizhevsky_NIPS2012}     			   & 0.722\\\midrule
		temporal$^\ast$ & 0.781 \\
		spatial$^\ast$  & 0.756 \\ 
		temporal \& spatial$^\ast$ & 0.784\\
		T with repetitions$^\ast$ & 0.794\\
		S\&T with repetitions$^\ast$ & \textbf{0.804} \\\midrule
		CliqueCNN$^\ast$ \cite{Bautista2016} 				& 0.790\\\bottomrule

	\end{tabular}
	%}
	\caption{Comparative posture analysis performance of auxiliary tasks in Olympic Sports dataset. Methods with $(\ast)$ are initialized with Imagenet pre-trained weights of Alexnet.}
	\label{table:postureOSD3}
\end{table}

% \FloatBarrier
\begin{figure*}
	\includegraphics[width=\linewidth]{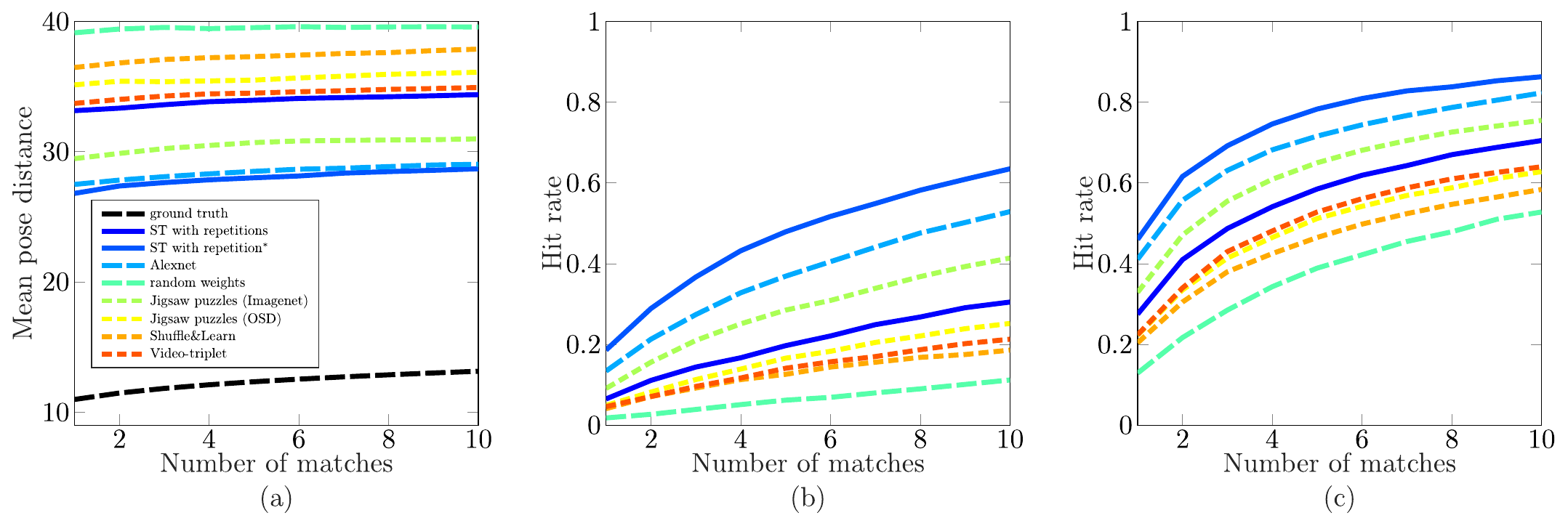}
	\caption{Pose retrieval results on MPII validation set: (a) Mean pose distance, (b) Hit rate\at K using nearest neighbor criterion, (c) Hit rate\at K using relative distance criterion. Model with ($\ast$) initialized with Imagenet pre-trained weights.}
	\label{fig:mpii_poseretrieval}
\end{figure*}
Second, we analyze the contributions of repetition mining and the two individual auxiliary tasks in \tabref{table:postureOSD3}. Temporal ordering underperforms with respect to spatial placement when initialized with random weights. We argue that temporal ordering is a more challenging task, since the temporal nature of actions has to be learned by the network. When the network is initialized from Imagenet, temporal ordering performs well. It has already learned to filter relevant visual information and improves with additional temporal cues from videos. On the other hand, spatial placement does not improve on Alexnet by such a large margin, because pre-trained Alexnet already comes with a good localization ability. In both settings (initialized randomly or from Imagenet pre-trained weights), repetition mining further boosts performance. This improvement highlights the benefit of the usage of repetitions.

Additionally in \tabref{table:postureOSD3}, we compare our best performing method with related work. When randomly initialized, our method performs better than several different self-supervised methods \cite{Doersch_ICCV2015,Wang_ICC2015,Norozzi_ECCV2016} and surpasses the best competitor by nearly 5 points. It even approaches the performance of the Imagenet pre-trained Alexnet, which is impressive considering that our training leveraged 680 sport videos (approx. 80K frames used) without labels, whereas Imagenet contains 1.2M labeled images. In the case of finetuning our model improves about 8 points on ImageNet pre-trained Alexnet and surpasses CliqueCNN \cite{Bautista2016} which is trained in the same setting.

\subsection{Pose Retrieval} \label{subsection:poseretrieval}
In this set of experiments we want to study the ability of our trained pose embeddings to generalize to unseen datasets. For this purpose, we evaluate our methods in the task of pose retrieval on the challenging MPII Human Pose dataset. We adopt the same procedure as described in \cite{Kwak_CVPR2016}, and split the fully annotated MPII training set into train and validation set. The validation set is further split in 1919 images for query and 8000 images for test purposes. The input images and pose annotations are normalized with respect to smallest square patch tightly enclosing all body part locations, and normalized into the input size of our network. 

According to \cite{Kwak_CVPR2016} three performance metrics are used: \emph{mean pose distance}, \emph{hit rate using nearest neighbor} and \emph{relative distance criterion}. The pose distance is the mean of Euclidean distances between normalized pose vectors. The mean pose distance is computed across the first K nearest neighbors. The hit rate measures the correctness of retrieval and is defined in two different ways: 1) \emph{nearest neighbor criterion} determines whether at least one retrieval among the K nearest neighbors belongs to the first fifty nearest neighbors in the pose space. 2) \emph{relative distance criterion} uses a $+10$ margin of minimum pose distance between query and test set.

The pose retrieval results evaluated on the three performance metrics on MPII are shown in \figref{fig:mpii_poseretrieval}. Here, we trained our method on the spatial\&temporal (ST) tasks with repetition mining using OSD only. It successfully generalizes to the challenging MPII dataset. When randomly initialized, it shows better mean pose distance and hit rate performance than previous methods, which are also trained on videos \cite{Wang_ICC2015,Misra_ECCV2016}. 

When the jigsaw puzzles method \cite{Norozzi_ECCV2016} is trained on the larger Imagenet dataset, they clearly outperform our method. We argue that this performance gap is due to different training data. To support this assumption, we re-train their method on OSD person boxes using their official implementation \footnote{https://github.com/MehdiNoroozi/JigsawPuzzleSolver}, and find it to perform worse than our self-supervised method across all measures. 

When initialized from Imagenet pre-trained weights, our method outperforms Alexnet across all measures particularly in hit rates. 

%Learning spatial and temporal relations jointly together with a motion-based curriculum and repetition mining makes our network aware of many useful relations for pose retrieval, and performs better than most state-of-the-art self-supervised approaches.

%\vspace{6pt}
\subsection{Pose Estimation}  \label{subsection:poseestimation}
For pose estimation we evaluate on the Leeds Sports Pose dataset \cite{Johnson2010}. We follow the procedure described in \cite{Bautista2016} and use the 1000 training images and 3938 (fully annotated) images from the extended training set as test set for retrieval while the original 1000 test images are used as query. In both query and test images, joint locations are normalized into our network’s input size.

\begin{table}[th!]
	\centering
	\resizebox{0.95\columnwidth}{!}{
		\begin{tabular}{*7l | l}    \toprule  % *8l
			\emph{Method} 						& \emph{Head} & \emph{Torso} & \emph{U.arms} & \emph{L.arms} & \emph{U.legs} & \emph{L.legs} & \emph{Mean} \\\midrule
			random weights             			& 19.3 & 45.2 & 9.6  & 4.1  & 21.1 & 20.3 & 19.9 \\
			ground truth 				 		& 72.4  & 93.7  & 58.7  & 36.4 & 78.8 & 74.9 & 69.2\\\midrule
			% Chen \& Yuille\cite{Chen_NIPS2014} 			& 92.7 & 87.8 & 69.2 & 55.4 & 82.9 & 77.0 & 75.0\\
			Chu \emph{et al.}\cite{Chu2016} 			& 89.6 & 95.4 & 76.9 & 65.2 & 87.6 & 83.2 & 81.1\\\midrule
			Shuffle\&Learn \cite{Misra_ECCV2016}  & 36.7 & 66.6 & 20.1 & 8.3  & 37.5 & 35.0 & 34.0 \\
			Video triplet \cite{Wang_ICC2015}     & 40.5 & 76.6 & 23.9 & 10.0 & 46.1 & 39.6 & 39.4 \\
			Jigsaw puzzles \cite{Norozzi_ECCV2016} (Imagenet)  & 49.3 & 80.1 & 27.5 & 11.9 & 50.5 & 47.4 & \textbf{44.4} \\
			Jigsaw puzzles \cite{Norozzi_ECCV2016} (OSD)       & 41.0 & 72.8 & 23.8 & 12.2 & 43.0 & 39.8 & 38.7\\
            S\&T with repetitions  				& 40.3  & 74.7  & 23.8  & 11.5  & 45.8  & 42.8 & 39.8 \\\midrule
			Alexnet \cite{Krizhevsky_NIPS2012}  & 42.4  & 76.9  & 47.8 & 41.8 & 26.7 & 11.2 & 41.1\\
			CliqueCNN \cite{Bautista2016} $^\ast$ 		& 45.5  & 80.1  & 27.2  & 12.6 & 50.1 & 45.7 & 43.5 \\
            S\&T with repetitions$^\ast$    	& 55.8  & 86.5  & 35.0  & 18.9  & 58.7  & 53.8 & \textbf{51.5}\\
			\hline
		\end{tabular}
	}
	\caption{Pose estimation results in Leeds Sports Pose dataset with PCP measures for each method. Methods with $(\ast)$ are initialized with Imagenet pre-trained weights.}
	\label{table:leeds_results}
\end{table}

\begin{figure}[b!]
	\begin{center}
		%\fbox{\rule{0pt}{2in} \rule{.9\linewidth}{0pt}}
		\includegraphics[width=.90\linewidth]{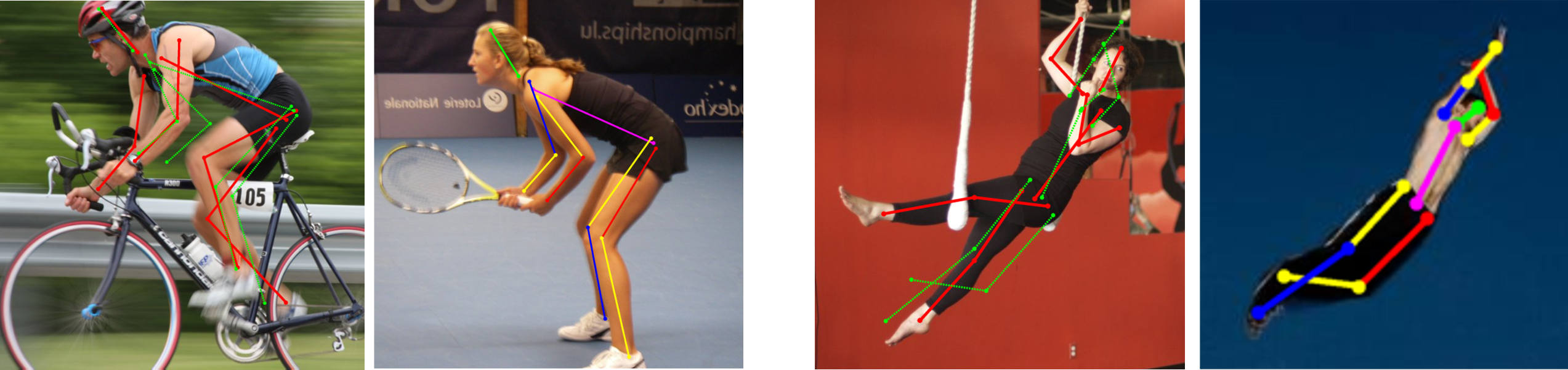}
	\end{center}
	
	\caption{Pose estimation results in Leeds Sports Pose dataset. First images are from test set with the superimposed ground truth skeleton depicted in {\color{red} red} and the predicted skeleton in {\color{green} green}. Second images are corresponding nearest neighbors.}
	%, which were used to transfer pose.
	\label{fig:posestimation}
\end{figure}

We report the Percentage of Correct Parts (PCP) measure \cite{Ferrari2008} on 14 body joints for different methods. According to PCP a part is considered correct, if its endpoints are within 50\% part length of the corresponding ground truth endpoints. 

Unsupervised pose estimation results of LSP in \tabref{table:leeds_results} show that our method, when initialized randomly, performs better than other self-supervised methods except for jigsaw puzzles trained on Imagenet. As in the case of pose retrieval, we argue that it is due to the size of Imagenet. When initialized from pre-trained weights, our method clearly outperforms \cite{Krizhevsky_NIPS2012,Bautista2016}.

Some qualitative samples from the query set together with their nearest neighbors are shown in \figref{fig:posestimation}. Our method is able to retrieve similar poses even if the query is very different from our training data.

In addition to our unsupervised experiments, we use our pose embeddings as an initialization of the supervised DeepPose \cite{Toshev_CVPR2014} method. In total, we evaluate four different initializations of \cite{Toshev_CVPR2014} on the MPII dataset: (i) our randomly initialized spatial\&temporal (ST) with repetitions model, (ii) Shuffle\&Learn \cite{Misra_ECCV2016}, (iii) random initialization, and (iv) Imagenet pre-trained Alexnet \cite{Krizhevsky_NIPS2012}.

\begin{table}[th!]
	\centering
	\resizebox{0.95\columnwidth}{!}{
		\begin{tabular}{*5l}    \toprule 
			
			& Ours 	& 	Shuffle\&Learn \cite{Misra_ECCV2016} 	& 	Random init.	&	Alexnet\cite{Krizhevsky_NIPS2012} \\ \midrule
			Head	 	 & 82.6		&	75.8				&	79.4			&	87.2							\\
			Neck	 	 & 90.3		&	86.3 	  			&	87.1			&	93.2							\\
			LR Shoulder	 & 79.5		&	75.0				&	71.6			&	85.2							\\
			LR Elbow	 & 62.8		&	59.2				&	52.1			&	69.6							\\
			LR Wrist	 & 47.1		&	42.2				&	34.6			&	52.0							\\
			LR Hip	 	 & 75.5		&	73.3				&	64.1			&	81.3							\\
			LR Knee	 	 & 65.3		&	63.1				&	58.3			&	69.7							\\
			LR Ankle	 & 59.5		&	51.7				&	51.2			&	62.0							\\
			Thorax		 & 90.1		&	87.1				&	85.5			&	93.4							\\
			Pelvis		 & 80.3		&	79.5				&	70.1			&	86.6							\\ \midrule
			Total		 & 73.3		&	69.3				&	65.4			&	78.0							\\
			\bottomrule
		\end{tabular}
	}
	\caption{PCKh@0.5 measure for DeepPose method \cite{Toshev_CVPR2014} on MPII Pose benchmark dataset comparing different initialization approaches.}
	\label{table:mpii_transfer}
\end{table}

For all initializations, we train the DeepPose method using the same setup and evaluate using PCKh@0.5 metric as shown in Table 4. Our method shows an improvement of 7.9\% and 4\% compared with random initialization and Shuffle\&Learn, respectively. It is only 4.7\% below Alexnet, which is learned using the labels of 1.2 million images.

\section{Conclusion}
In this paper, we have proposed two complementary self-supervised tasks, temporal ordering and spatial placement which are trained jointly on unlabeled video data. To boost self-supervised training, we have introduced a motion-based curriculum and a procedure for mining repetitive poses and using them as valuable training data. Our pose embeddings capture the characteristics of human posture, which we have demonstrated in experiments on pose analysis. In the Olympics Sports dataset, the learned representation decreases the gap between self-supervised methods and Imagenet supervision, and fine-tuning with our self-supervised approach significantly improves the performance of models pre-trained on Imagenet. Finally, we have shown that the trained embeddings are able to generalize to unseen datasets in pose analysis without fine-tuning. 

%Our method as an initialization to DeepPose also proves that it is able to capture relevant relations for pose understanding and can be used to learn better supervised methods with the same amount of data.

\paragraph{Acknowledgments:} This work has been supported in part by the Heidelberg Academy for the Sciences, DFG, and by an NVIDIA hardware grant.

%\newpage

{\small
\bibliographystyle{ieee}
\bibliography{egbib}
}

\end{document}